\pdfoutput=1

\documentclass[11pt]{article}

\usepackage[final]{acl}

\usepackage{times}
\usepackage{latexsym}

\usepackage[T1]{fontenc}

\usepackage[utf8]{inputenc}

\usepackage{microtype}

\usepackage{inconsolata}

\usepackage{graphicx}
\usepackage{xcolor}
\usepackage{pifont} 
\usepackage{amsmath}
\usepackage{amsthm}
\usepackage{amssymb}
\usepackage{booktabs}
\usepackage{enumitem}

\definecolor{myRed}{rgb}{0.808,0.067,0.149}
\definecolor{myGreen}{rgb}{0.067,0.708,0.149}
\newcommand{\xmark}{{\color{myRed}\ding{55}}}%
\newcommand{\cmark}{{\color{myGreen}\ding{51}}} 
\patchcmd{\abstract}{-5.5em}{0em}{}{}

\title{Text Classification Under Class Distribution Shift: A Survey}

\author{
\textbf{Adriana-Valentina Costache}$^{1,*}$\textbf{, Silviu-Florin Gheorghe}$^{1,*}$\textbf{, Eduard Poesina}$^{1,*}$\textbf{,}\\
\textbf{Paul Irofti}$^1$\textbf{, Radu Tudor Ionescu}$^{1,\diamond}$\\
$^1$Department of Computer Science, University of Bucharest, Romania\\
$^{*}$Equal contribution. $^{\diamond}${\texttt{raducu.ionescu@gmail.com}}
}

\begin{document}
\maketitle

\begin{abstract}
The basic underlying assumption of machine learning (ML) models is that the training and test data are sampled from the same distribution. However, in daily practice, this assumption is often broken, i.e.~the distribution of the test data changes over time, which hinders the application of conventional ML models. One domain where the distribution shift naturally occurs is text classification, since people always find new topics to discuss. To this end, we survey research articles studying open-set text classification and related tasks. We divide the methods in this area based on the constraints that define the kind of distribution shift and the corresponding problem formulation, i.e.~learning with the Universum, zero-shot learning, and open-set learning. We next discuss the predominant mitigation approaches for each problem setup. We further identify several future work directions, aiming to push the boundaries beyond the state of the art. Finally, we explain how continual learning can solve many of the issues caused by the shifting class distribution. We maintain a list of relevant papers at \url{https://github.com/Eduard6421/Open-Set-Survey}.
\end{abstract}

\setlength{\abovedisplayskip}{3.5pt}
\setlength{\belowdisplayskip}{3.5pt}

\vspace{-0.15cm}
\section{Introduction}
\vspace{-0.15cm}

The primary assumption of any machine learning (ML) model is that the data is independent and identically distributed (IID) \citep{Vapnik-SB-1995}. In learning theory, the IID assumption plays a critical role, as it guarantees the generalization capacity of models, provided that sufficient training data is available, and that the hypothesis class is not too large. However, this assumption does not always hold in daily practice. One such example is text categorization by topic, where new topics naturally emerge as the interest of people changes over time. For example, journalists started publishing news articles about the COVID-19 pandemic only after its outbreak began in December 2019\footnote{\url{https://en.wikipedia.org/wiki/COVID-19_pandemic}}. Hence, if a classifier is trained in a closed-world setup (a scenario where it is presumed that every instance encountered by the model belongs to a class that is present in the training data), there is a high risk that the respective model will encounter instances from classes that were not present in the training data, during operation. Such instances will be misclassified as belonging to one of the training classes, degrading the overall performance of the ML system, and implicitly reducing user satisfaction.

\begin{figure}[t]
\centering 
\includegraphics[width=1.0\linewidth]{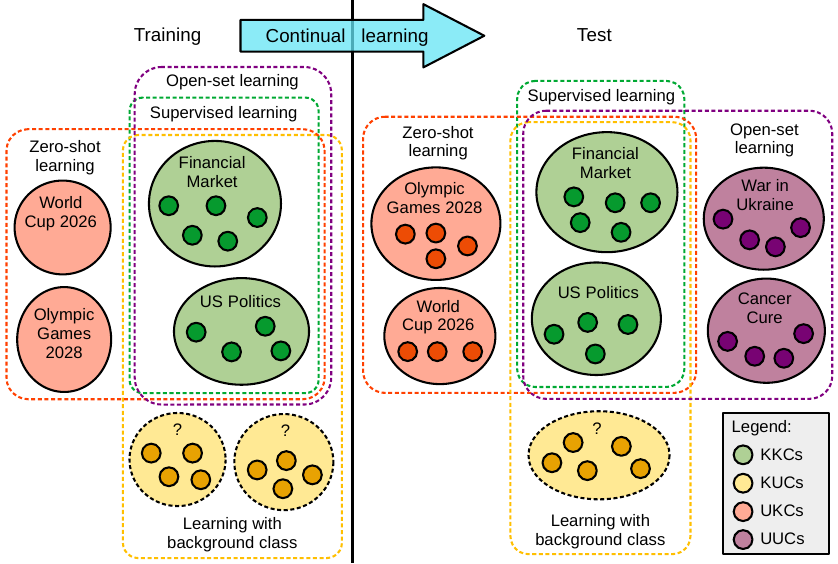} 
\vspace{-0.65cm}
\caption{An illustration of text classification by topic under class distribution shift, resulting in four class categories and associated learning problems. Known Known Classes (KKCs) correspond to standard supervised learning, Known Unknown Classes (KUCs) to learning with background class, Unknown Known Classes (UKCs) to zero-shot learning, and Unknown Unknown Classes (UUCs) to open-set learning. Continual learning provides viable solutions to mitigate class distribution shift. Best viewed in color.}
\label{fig:teaser}
\vspace{-0.3cm}
\end{figure}

\begin{table*}[t!]
    \centering
    \small{
    \begin{tabular}{p{0.09\textwidth} p{0.27\textwidth} p{0.20\textwidth} p{0.32\textwidth}}
        \toprule
        \textbf{Category} & \textbf{Definition} & \textbf{Problem \mbox{Formulation}} & \textbf{Representative Studies} \\
        \midrule
        KKCs  
        & Classes for which we have labeled training data. 
        & Standard supervised learning & (out of scope)\\
        \midrule
        KUCs 
        & Classes with available training examples, but for which there are no class labels. 
        & Classification with background / Universum class & \citep{Dhamija-NeurIPS-2018,Hendrycks-ICLR-2019,Lee-ICLR-2018,Liu-NeurIPS-2020}  \\
        \midrule
        UKCs 
        & Classes that are known to exist, but for which we have no examples during training.
        & Zero-shot learning & \citep{Yin-EMNLP-2019,Gera-EMNLP-2019,Meng-EMNLP-2020,Meng-NIPS-2022,Sanh-ICLR-2022,Zhang-ARXIV-2024} \\
        \midrule
        UUCs 
        & Classes whose existence is unknown during training and for which there is no training data. 
        & Open-set learning (and discovery) & \citep{Chen-KDD-2023,Walkowiak-IntelliSys-2020,Chen-ACL-2024,Kim-BigData-2022,Walkowiak-IEA-2019} \\
        \bottomrule
    \end{tabular}
    }
        \vspace{-0.25cm}
    \caption{Types of classes that can naturally emerge in classification under class distribution shift. For each class category of interest, we provide the corresponding problem formulation as well as a set of representative works from the NLP or ML domains.}.
    \label{tab:class-types}
    \vspace{-0.5cm}
\end{table*}

A large body of works, e.g.~\citep{Geng-TPAMI-2020,Salehi-TMLR-2022,Vaze-ICLR-2021,Yang-IJCV-2024}, studies the adaptation and application of ML models when the class distribution changes from training time to inference time. One of the first articles to highlight this problem is that of  \citet{Scheirer-TPAMI-2012}. The authors define the \emph{open-set risk} as the sum of the empirical risk and the open-space risk, the latter being defined as the risk of misclassifying instances of unknown classes as belonging to a known class. Over time, researchers attempted to minimize the open-set risk by addressing a broad range of formulations of the class distribution shift problem. These formulations can be mainly structured according to the class categories proposed by \citet{Geng-TPAMI-2020}. 

We present the class categories and the corresponding problem formulations in Figure \ref{fig:teaser}. We further define the class categories and provide representative studies for each problem formulation in Table \ref{tab:class-types}. KKCs lead to the conventional supervised learning paradigm, which is well-studied and results often surpass human-level performance \citep{Cozma-ACL-2018,Tedeschi-ACL-2023}. Frameworks that deal with KUCs typically use a background (a.k.a.~Universum) class \citep{Dhamija-NeurIPS-2018,Weston-ICML-2006}, essentially extending the supervised learning setup with a new category where all the examples that do not belong to any of the KKCs are placed. The zero-shot learning paradigm \citep{Yin-EMNLP-2019,Chaudhary-ESWA-2024,Pourpanah-TPAMI-2022} aims to deal with UKCs, namely with classes that are known in advance (at training time), but for which there are no data samples. Zero-shot learning aims to classify data samples into UKCs, but the main limitation is that the set of UKCs is fixed. As a consequence, dealing with UUCs is not possible in the zero-shot learning setup. Open-set learning \citep{Geng-TPAMI-2020,Scheirer-TPAMI-2012} aims to partially address this challenge by identifying UUCs without having prior information about such classes during training. Sometimes, UUCs are identified via a two-stage pipeline that combines an outlier detection (a.k.a.~novelty detection) model \citep{Chen-KDD-2023,Walkowiak-IntelliSys-2020,Walkowiak-ASMBI-2018} and a supervised model trained on KKCs. However, unlike zero-shot learning methods, open-set methods do not aim to classify the data samples into UUCs. A more comprehensive setup is proposed in \citep{Zheng-CVPR-2022}, where the authors aim not only to detect samples belonging to UUCs, but also to classify them. This framework, called \emph{open-set learning and discovery}, can be seen as a generalization of zero-shot learning, where the set of UKCs can be updated during inference, therefore transforming it into a set of UUCs.

To date, open-set classification and the related tasks discussed above were mostly studied in the vision domain, where various tasks have been explored, such as object recognition \citep{Vaze-ICLR-2021,Scheirer-TPAMI-2012,Kong-ICCV-2021}, semantic segmentation \citep{Cen-ICCV-2021,Oliveira-ML-2023}, object detection \citep{Zheng-CVPR-2022,Dhamija-WACV-2020,Liu-ECCV-2023}, and video anomaly detection \citep{Acsintoae-CVPR-2022,Wu-CVPR-2024}, among others. Comparatively less attention has been dedicated to this family of tasks in the text domain \citep{Chen-ACL-2024,Kim-BigData-2022,Fei-NAACL-2016}. However, class distribution shift is a prevalent phenomenon in text classification. Aside from the example provided earlier about the changing topics over time, there are many other natural language processing (NLP) tasks where the class distribution can shift. For example, in authorship identification, new authors can emerge over time, so an authorship identification model needs to update the list of potential authors. Another task affected by class distribution shift is intent detection in conversational AI. For instance, a chat bot trained to recognize intents such as ``book flight'' or ``check weather'' may encounter new intents such as ``play music'' at test time. Similarly, in named entity recognition, a system trained to recognize entities such as ``person'' and ``location'' might face new entity types such as ``event'' or ``disease'' during operation. Despite the clear likelihood of encountering new classes during inference in various NLP tasks, to the best of our knowledge, there is no survey on open-set text classification. To this end, we review articles that address the class distribution shift in the text domain. We further propose future work directions to address the challenges of open-set learning and discovery in NLP tasks, most of which stem from the continual learning paradigm \citep{Wang-TPAMI-2024}.

It is perhaps important to note that class distribution shift is a particular kind of dataset shift. The dataset shift problem was studied by  \citet{Moreno-PR-2012}, who meticulously categorize the various types of dataset shift. However, the authors did not specifically include class distribution shift as an interesting scenario, but they recognize the existence of dataset shift scenarios that \textit{``are so hard that we currently consider them impossible to solve''} \citep{Moreno-PR-2012}. Since 2012, the state of research in machine learning has changed significantly, and many problems that were thought as being impossible to solve are now actively studied. For instance, there are several recent surveys, e.g.~\citep{Geng-TPAMI-2020,Parmar-CSUR-2023,Zhu-Arxiv-2024}, that focus on the open-world learning problem, where new classes can emerge during testing. Yet, most of the studies covered by these surveys come from computer vision and image processing. To the best of our knowledge, we are the first to focus specifically on class distribution shift in natural language processing. 

In summary, our contribution is twofold:
\begin{itemize}[leftmargin=*]
    \item \vspace{-0.2cm} We provide a literature review of open-set text classification and related tasks, including zero-shot text classification and learning with the Universum class.
    \item \vspace{-0.2cm} We propose several future work directions for open-set learning and discovery, a task that has not been extensively explored in NLP.
\end{itemize}



\vspace{-0.15cm}
\section{Notations and Definitions}
\vspace{-0.15cm}

\noindent
\textbf{Notations.}
We use the following notations to define the various task formulations covered in our survey. Let $x \in \mathcal{X}$ represent a text sample from the data space $\mathcal{X}$, and $y \in \mathcal{Y}$ a label from the label space $\mathcal{Y}$. The label space is divided into four disjoint subsets denoted as known known classes ($C^{kk}$), known unknown classes ($C^{ku}$), unknown known classes ($C^{uk}$), and unknown unknown classes ($C^{uu}$), such that $C^{kk} \cup C^{ku} \cup C^{uk} \cup C^{kk}= \mathcal{Y}$ and the intersection between any two subsets is the empty set, e.g.~$C^{kk} \cap C^{uk} = \emptyset$.


\noindent
\textbf{Learning with background class.}
Learning with background / Universum class aims to detect samples from $C^{ku}$, while having samples from both $C^{kk}$ and $C^{ku}$. Formally, the training data is defined as $\mathcal{D} = \{(x, y) \mid x \in \mathcal{X}, y \in C^{kk} \cup C^{ku} \}$, while the test data is defined as $\mathcal{T} = \{(x, y) \mid x \in \mathcal{X}, y \in C^{kk} \cup C^{ku}\}$. In this setup, a classifier is defined as $h: \mathcal{X} \to C^{kk} \cup \{\text{background}\}$, i.e.~for any sample that belongs to $C^{ku}$, the model should label the respective sample as background.

\noindent
\textbf{Zero-shot text classification.}
Zero-shot learning aims to recognize samples from $C^{uk}$ using knowledge transferred from $C^{kk}$. Formally, the training data is defined as $\mathcal{D} = \{(x, y) \mid x \in \mathcal{X}, y \in C^{kk}\}$, while the test data is defined as set $\mathcal{T} = \{(x, y) \mid x \in \mathcal{X}, y \in C^{uk}\}$. Knowledge is usually transferred via an auxiliary semantic space $A$, e.g.~word embeddings or attribute vectors, such that $f: \mathcal{X} \to A$ and $g: A \to \mathcal{Y}$. The mapping $f$ projects features into the semantic space, while $g$ maps semantic representations to class labels.

\noindent
\textbf{Open-set text classification and discovery.} Open-set learning aims to classify samples from $C^{kk}$, while rejecting samples from $C^{uu}$ encountered during inference. The training set is defined as $\mathcal{D} = \{(x, y) \mid x \in \mathcal{X}, y \in C^{kk}\}$, while the test set is given by $\mathcal{T} = \{(x, y) \mid x \in \mathcal{X}, y \in C^{kk} \cup C^{uu}\}$. An open-set learning model is defined as $h: \mathcal{X} \to C^{kk} \cup \{\text{unknown}\}$, i.e.~for any sample that belongs to $C^{uu}$, the model should label the respective sample as unknown. 
In the \emph{open-set learning and discovery} setup, the model is defined as $h: \mathcal{X} \to C^{kk} \cup C^{uu}$, i.e.~the model must be able to classify any data sample, regardless of the fact that the sample belongs to $C^{kk}$ or $C^{uu}$. To classify samples into $C^{uu}$, classes belonging to $C^{uu}$ must be discovered during inference.

\vspace{-0.15cm}
\section{Optimization Objectives}
\vspace{-0.15cm}

Starting from a model whose parameters $\theta$ are obtained by training on a given dataset $\mathcal{X}$ with labels $\mathcal{Y}$,
our general optimization problem is:
\begin{equation}
\begin{split}
\theta^\star &=
  \arg\min_{\theta} \mathcal{L}\left(\theta, \mathcal{X}, \mathcal{Y}\right)\\
  &=
  \arg\min_{\theta} \mathbb{E}_{(x,y) \sim(\mathcal{X}, \mathcal{Y})} \left[  f_\theta(x,y) \right].
\end{split}
\end{equation}
We recover below the particular learning frameworks by giving specific loss formulations w.r.t.~the class types defined in Table~\ref{tab:class-types}. The supervised cross-entropy on KKCs is:
\begin{equation}
\mathcal{L}_{\text{KKC}}(\theta,\mathcal{X},C^{kk})\!\!=\!\! \mathbb{E}_{(x,y)\sim(\mathcal{X},C^{kk})}\!\left[-\!\log p_\theta(y \!\mid\! x)\right]\!.
\end{equation}
The binary cross-entropy on KKCs vs.~KUCs is:
\begin{equation}
\begin{split}
\!\!\!\mathcal{L}_{\text{KUC}}& (\theta,\mathcal{X},C^{kk}\cup C^{ku})\!\!=\!\! \mathbb{E}_{(x,y)\sim(\mathcal{X},C^{kk}\cup C^{ku})}\\
& \left[-y\!\cdot\!\log f_\theta(x)\!+\!(1\!-\!y)\!\cdot\!\log(1\!-\!f_\theta(x)) \right]\!.
\end{split}
\end{equation}
The contrastive learning objective on UKCs is:
\begin{equation}
\begin{split}
\mathcal{L}_{\text{UKC}}&(\theta,\mathcal{X},C^{uk})\!=\!\mathbb{E}_{(x,y)\sim\mathcal (\mathcal{X}, C^{uk})}\\
& \left[-\!\log \frac {\exp \left( \text{sim} (f_\theta(x), s_y)\right)}{\sum_{c\in\mathcal C^{uk}} \exp \left( \text{sim} (f_\theta(x),s_c) \right)} \right]\!,
\end{split}
\end{equation}
where $s_y$ is a prototype vector (embedding) for class $y$, which can be obtained by passing the class names through the model $f_\theta$, and $\text{sim}$ is a similarity measure (e.g.~the cosine similarity).
Finally, the supervised cross-entropy on UUCs is:
\begin{equation} 
\mathcal{L}_{\text{UUC}}(\theta,\mathcal{X},C^{uu})\!\!=\!\! \mathbb{E}_{(x,y)\sim(\mathcal{X},C^{uu})}\!\left[-\!\log p_\theta(y\!\mid\!x)\right]\!, 
\end{equation}
and the label set is computed as $C^{uu}\!=\!\{y \mid y\!=\! g_\omega (x'), \forall x' \in \mathcal{I}(h_\phi (x)) \}$,
where
$h_\phi$ is an outlier detection method (e.g.~Isolation Forrest)
and
$\mathcal{I}$ is an unsupervised outlier indicator function
that selects outlier samples $x'$, which are further given as input to a clustering method $g_\omega$ (e.g.~k-means).

Further, we can integrate all learning frameworks into a joint optimization objective:
\begin{equation}
\mathcal{L}_{\text{joint}}\!=\!\mathcal{L}_{\text{KKC}}\!+\!\alpha\!\cdot\! \mathcal{L}_{\text{KUC}}\!+\!\beta\!\cdot\!\mathcal{L}_{\text{UKC}}\!+\!\gamma\!\cdot\! \mathcal{L}_{\text{UUC}}, 
\end{equation}
where $\alpha$, $\beta$ and $\gamma$ are scalar values that can be used to control the optimization within a particular learning framework, as follows:
\begin{itemize}[leftmargin=*]
\item \vspace{-0.2cm} Learning with background class: $\alpha\!>\!0$, $\beta\!=\!0$, $\gamma\!=\!0$.
\item \vspace{-0.2cm} Zero-shot learning: $\alpha\!=\!0$, $\beta\!>\!0$, $\gamma\!=\!0$.
\item \vspace{-0.2cm} Open-set learning and discovery: $\alpha\!=\!0$, $\beta\!=\!0$, $\gamma\!>\!0$.
\end{itemize}

\vspace{-0.15cm}
\section{Learning with Background Class}
\vspace{-0.15cm}


The problem of learning with a background class has been addressed through proxy tasks such as out-of-distribution (OOD) detection or outlier rejection \citep{Dhamija-NeurIPS-2018}. These approaches improve the reliability of the model by managing uncertainty and rejecting inputs outside the distribution of KKCs. Research in this area has been predominantly driven by computer vision studies, whereas text classification remains comparatively underexplored. However, many techniques can be easily adapted to NLP tasks by replacing image-based feature extractors with robust language encoders.

A straightforward approach to detect background samples involves training a classifier to refine the decision boundary between the original dataset and a curated background dataset. This technique is commonly employed in computer vision tasks \citep{Mullapudi-IEEE-2021}, leveraging the abundance of background images available, e.g.~iNaturalist-BG, Places-BG, etc.

Another common approach is to employ custom regularization techniques for background-class learning. For Support Vector Machines (SVMs), \citet{Weston-ICML-2006} introduce a penalty term that incorporates the distance between the decision boundary and background-class (Universum-class) data points. Their method is based on the principle that Universum-class samples should lie near the decision boundary, reflecting real-world uncertainty and ambiguity in classification. This regularization technique enforces a decision-making process that accounts for inherent uncertainty in the data. Expanding this idea, \citet{Zhou-EMNLP-2023} propose a learning method that distinctively treats the classes of interest and the Universum class. They introduce closed decision boundaries for the classes of interest and define the space outside these boundaries as belonging to the Universum. To determine whether a sample belongs to a class of interest or to 
the Universum class, the authors propose a probability estimation based on inter-class rules. This approach assesses whether a sample lies predominantly within the closed decision boundary associated with a class of interest. If the sample does not clearly fit into one of the classes, it is considered to belong to the Universum. \citet{Dhamija-NeurIPS-2018} introduce a more generic regularization based on two loss functions. The Entropic Open-Set Loss aims to reduce entropy for KKCs, while maximizing entropy for KUCs, ensuring that the model treats unfamiliar data with the highest possible uncertainty. The Objectosphere Loss strengthens the separation between KKCs and KUCs by increasing the feature magnitudes for KKC samples, while reducing them for KUCs. These loss functions create a more representative feature space, improving the ability of the model to reject out-of-distribution inputs. 

Developing novel loss functions represents a promising line of work, considered by other researchers as well. For instance, \citet{Liu-NeurIPS-2020} propose energy-based models as a viable solution for background sample detection, where the main contribution is a loss function designed to include both a standard classification loss and a model-specific penalty term. This additional term penalizes high-energy representations of KKC samples, while suppressing low-energy assignments for KUC samples, ensuring a clear energy-based separation between the two categories. \citet{Hendrycks-ICLR-2019} introduce the concept of outlier exposure, which involves training models on an auxiliary dataset containing OOD samples. Their objective function combines the original classification loss with an additional penalty that makes the model assign low confidence to out-of-distribution inputs. For softmax-based classification tasks, this approach encourages predictions for outliers to follow a uniform distribution. In density estimation tasks, the outlier exposure loss often incorporates a margin ranking objective, ensuring that in-distribution samples consistently receive higher log probabilities than outliers. \citet{Lee-ICLR-2018} propose a method to train neural networks for better OOD detection, while maintaining their classification accuracy. The contributions refer to introducing a confidence loss, leveraging a generative adversarial network (GAN) \citep{Goodfellow-NIPS-2014} for OOD sample generation, and combining these into a joint training framework. The confidence loss combines the standard cross-entropy loss with a KL-divergence term penalizing confident predictions for OOD samples, which are compared with a uniform distribution. The OOD samples are generated by a GAN, such that the samples lie close to the boundary of the in-distribution data. The authors train the classifier and the GAN jointly. A complementary direction is explored by \citet{Hu-ACM-2021}, who develop a loss function specifically designed to calibrate uncertainty in both in-distribution and OOD settings. Their framework employs evidential neural networks and adds two regularization terms: one penalizes high uncertainty for in-distribution samples, and the other rewards maintaining high uncertainty OOD samples. By explicitly modeling and balancing predictive confidence, their approach improves classification reliability under class distribution shift.


\vspace{-0.15cm}
\section{Zero-Shot Text Classification}
\vspace{-0.15cm}
When no examples from the target classes are available, various alternative tasks can be leveraged, with or without additional training, to perform zero-shot text classification. For instance, \citet{Yin-EMNLP-2019} propose using the entailment task as a zero-shot text classification task. To decide if a text $x$ should be classified into a certain class, e.g.~``politics'', one can ask if $x$ entails ``The previous text is about politics''. This procedure is applied for each of the possible classes, and the one with the best confidence is considered the correct one. Using this technique, any model capable of solving the entailment problem can implicitly be used to perform zero-shot text classification. The zero-shot approach based on entailment~\citep{Yin-EMNLP-2019} represents the basis for how large language models (LLMs) are used via prompting to solve various downstream tasks. LLMs \citep{Bommasani-ARXIV-2021,Yang-CSUR-2025, Zhou-IJMLC-2024} are usually pre-trained on a very large corpus of unlabeled data, using a variety of self-supervised tasks, e.g.~next token prediction, sentence order prediction, etc. During pre-training, LLMs learn general facts, such as language structure, making them ideal for task adaptation. LLMs are subsequently used in different downstream tasks, which often involve zero-shot setups. There are two main methodologies for adapting LLMs to a target task: \emph{fine-tuning} and \emph{instruction tuning}. 

Fine-tuning involves adapting a model to a task by performing an extra training step using task-specific data. Instruction tuning, defined as fine-tuning language models on a collection of datasets described via instructions, can be seen as a generalization over the framework proposed by \citet{Yin-EMNLP-2019}. Instruction tuning can lead to better performance on unseen tasks, if the target task is described via simple instructions provided in natural language. Models trained with instruction tuning can perform various tasks, including zero-shot classification, as shown by \citet{Zhang-ARXIV-2024}. This capability is achieved by training language models to respond to simple natural language commands, with some degree of generality. For instance, \citet{Wei-ICLR-2022} and \citet{Sanh-ICLR-2022} explore similar ideas, dividing the tasks for which datasets are available into separate clusters, each containing multiple datasets. Fine-tuning via instructions on any of the clusters leads to performance gains when the model is tested on tasks from the other clusters. \citet{Xu-EMNLP-2022} continue this line of work by increasing the number of tasks used for pre-training from a few dozen to over 1000, showing that increasing the number of tasks is a good alternative to increasing the model size. 

\begin{figure}[t!]
\centering
    \includegraphics[width=1.0\linewidth]{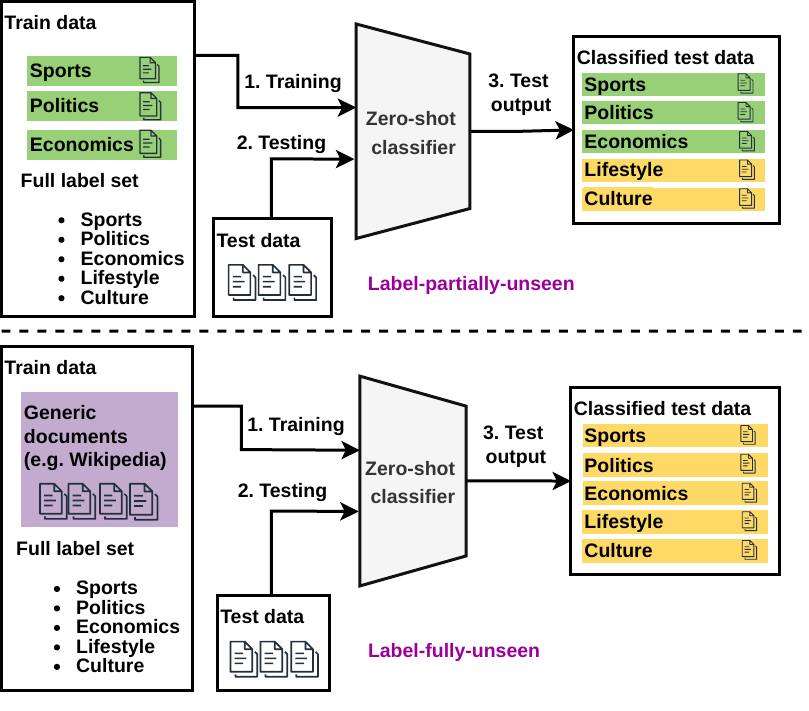} 
    \vspace{-0.75cm}
\caption{Zero-shot classification variants, as described by \citet{Yin-EMNLP-2019}. In \emph{label-partially-unseen}, the model is trained on labeled data from a subset of classes, then tested on the full set. In \emph{label-fully-unseen}, the model is trained on unlabeled data, possibly unrelated to the target task. Best viewed in color.}
\label{fig:ZS-Variants}
\vspace{-0.4cm}
\end{figure}

Another possible approach for zero-shot classification is to use an LLM to generate a synthetic dataset $\mathcal{S}$, tailored for the target task. Further, $\mathcal{S}$ can be used as training data in a fully-supervised setup, either to fine-tune an LLM or to train a conventional model from scratch. Although the dataset $\mathcal{S}$ can be made arbitrarily large, it can quickly hit diminishing returns, as it tends to contain redundant data \citep{Ye-EMNLP-2022}. Moreover, the performance obtained by using $\mathcal{S}$ is not as high as it would be when using a real human-labeled dataset of comparable size. To this end, several efforts have been made to improve the quality of $\mathcal{S}$ \citep{Meng-NIPS-2022,Ye-EMNLP-2022}. A generic dataset, such as Wikipedia, is sometimes employed as a source of additional diversity. \citet{Meng-NIPS-2022} use controlled text generation \citep{Hu-ICML-2018} to direct an LLM to generate texts that are relevant for a specific class label. To increase sample diversity, repeating sequences are penalized. Alternatively, \citet{Yu-ACL-2023} use Wikipedia as a general purpose corpus. In their method, called ReGen, a retrieval model is trained using contrastive learning for the task of finding the most relevant documents from the corpus. ProGen \citep{Ye-EMNLP-2022} and ZeroGen \citep{Ye-EMNLP-2022} represent other alternatives to generate synthetic datasets. ZeroGen uses prompt engineering to generate $\mathcal{S}$ from scratch, without the need of additional data. ProGen employs a feedback loop to iteratively refine a task-specific model.

Some studies relax the zero-shot learning setup in various ways, such as reducing the zero-shot setup to a subset of classes, while providing labeled examples for the others, or considering \emph{the availability of an unlabeled set of samples} from the target classes, but without knowing the corresponding class labels (see Figure \ref{fig:ZS-Variants}). For example, \citet{Meng-EMNLP-2020} train a model on a subset originating from the same distribution as the dataset used for evaluation, but without labels. This can be a valid setup, depending on the specific problem at hand, mainly when data is available, but labeling is difficult. \citet{Gera-EMNLP-2019} adopt a similar approach, based on self-training and entailment, for an instruction model. A model $h_0$ assigns pseudo-labels based on the confidence of entailment with ``This text is about $<$\texttt{class}$>$''. The resulting candidates are filtered and  used to train a new model $h_1$ that is further used to assign new pseudo-labels, and so on. Although this method may look similar with the one proposed by \citet{Yin-EMNLP-2019}, \citet{Gera-EMNLP-2019} use the entailment task to assign pseudo-labels to a dataset used for downstream training of a target model, while \citet{Yin-EMNLP-2019} directly employ entailment for classification.

\vspace{-0.15cm}
\section{Open-Set Text Classification and Discovery}
\vspace{-0.15cm}
\noindent
\textbf{Trends and approaches.}
A simple way to address the open-set learning problem is to classify the data using a closed-set algorithm, then estimate how well each sample corresponds to the assigned class. If the confidence (or similarity) is under a certain threshold, the example is considered to belong to the open space. In practice, this threshold can be difficult to set, because, in general, the ratio of instances belonging to $C^{uu}$ in the test set $\mathcal{T}$ cannot be estimated without prior knowledge. This problem is already identified and discussed by  \citet{Scheirer-TPAMI-2012}, who frame the issue as an effort to harmonize the empirical risk, measured on the training data with the unmeasurable \textit{open-space risk}, which is unknown. 

There are a few methods that avoid estimating this threshold. Starting from the observation that there are spaces where feature vectors of instances from the same class generally reside close together, some researchers use methods specific to clustering. In this context, an outlierness factor can be calculated to decide if an instance should be classified into a class or allocated to the open space. If an example is an outlier, it probably belongs to the open space.  Various outlierness factors have been proposed so far. \citet{Walkowiak-IEA-2019} employ the Local Outlier Factor (LOF), which uses a weighted Euclidean distance to find outliers based on the distance to their local neighbors. Subsequent studies of the same authors \citep{Walkowiak-IntelliSys-2020,Walkowiak-ICAISC-2019} use LOF as well. Another factor, called Angle-Based Outlier Factor, which was originally introduced by \citet{Kriegel-ACMSIGKDD-2008}, is later adapted for text classification by \citet{Walkowiak-ASMBI-2018}, under the name ABOF2.

Another solution to avoid setting an arbitrary threshold is to identify the examples belonging to the open space before performing the classification task. The problem is therefore broken into two subtasks, outlier detection (OD) and closed-set classification, which can be treated independently. In this setup, the classification part is closed, hence OD is the difficult part. 

\noindent
\textbf{Open-set solutions via outlier / novelty detection.}
\citet{Kannan-SDM-2017} divide the outlier detection algorithms for text into three subcategories: distance-based, density-based, and subspace-based. More recently, energy-based models have also been explored as a promising method for detecting out-of-distribution data. \citet{Grathwol-ICLR-2018} propose a Joint Energy-Based Model (JEM) by reinterpreting a softmax-based classification network as a generative model. This approach uses the logits to define an energy-based representation of the joint distribution of data points and labels. Another notable contribution to outlier detection is the typicality test, introduced by \citet{Nalisnick-ARXIV-2019}. This method determines whether a given data point belongs to the model’s typical set, rather than relying solely on likelihood estimation. The test is based on the observation that deep generative models can assign higher likelihoods to OOD data than to in-distribution samples. By focusing on typicality instead of raw probability density, this method offers a more robust approach to detecting outliers.


We emphasize that there are two main types of distribution shifts \citep{Baran-ACL-2023}: \emph{semantic shift}, defined as the occurrence of new classes, and \emph{background shift}, defined as changes that are class-agnostic, such as the level of formality, stance, and so on.  
If the OD detects background shift, the  examples belonging to $C^{kk}$ that exhibit background shift will not even make it to the classifier.   Consequently, the outlier detector should only detect semantic shift, but this is difficult without access to outliers during training. The subject of semantic vs.~background shift in text is discussed in detail by \citet{Arora-EMNLP-2021}. While outlier detection in general is very well studied \citep{Wang-ACCESS-2019}, there are very few articles focused on text, especially ones trying to isolate semantic shift. Yet, a simple way to detect semantic shift is to use rare terms \citep{Mohotti-TKDD-2020}.


\begin{figure}[t!]
\centering 
    \includegraphics[width=1.0\linewidth]{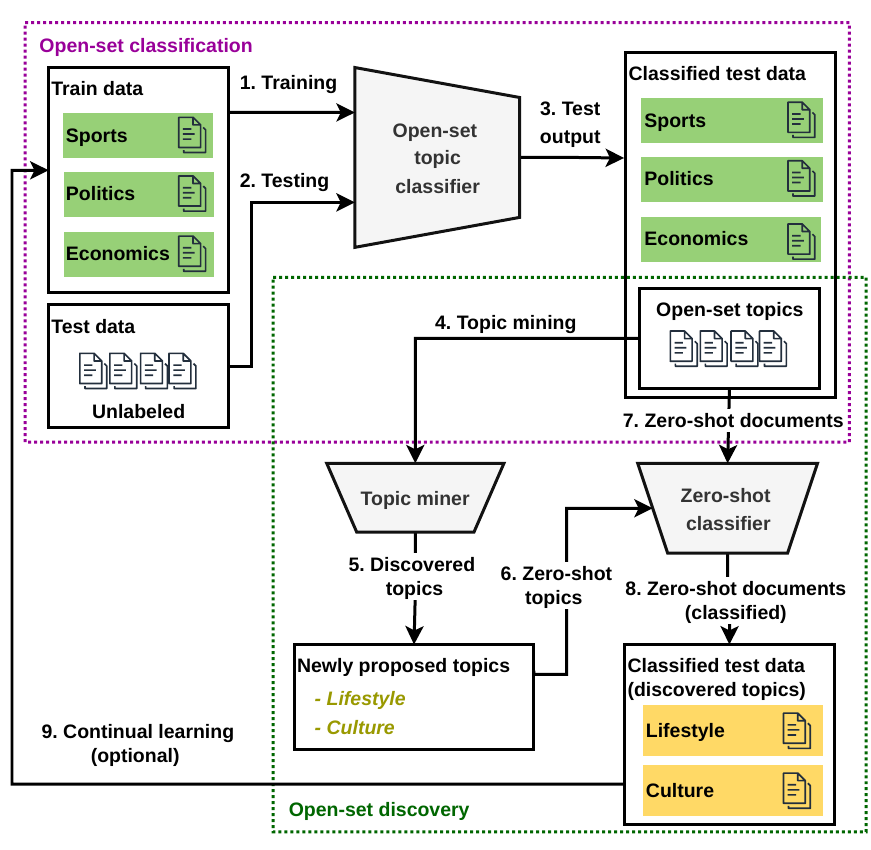} 
\vspace{-0.75cm}
\caption{A generic open-set classification and discovery pipeline. The data is initially classified by the open-set classifier. The open-space instances are mined for new class candidates. Then, a zero-shot classifier is used to find the relevant instances. The process can be reiterated via continual learning, including the newly found classes and corresponding documents to retrain the original open-set model. Best viewed in color.}
\label{fig:OSD-Arch}
\vspace{-0.4cm}
\end{figure}

\noindent
\textbf{Open-set semi-supervised text classification.}
Semi-supervised text classification (STC) has its own open-set variant, called open-set semi-supervised text classification (OSTC), introduced by \citet{Chen-KDD-2023}. For the STC task, the training set consists of some labeled examples from each class, and many additional unlabeled examples. In this setup, all the unlabeled examples are presumed to belong to $C^{kk}$. This assumption is difficult to enforce in practice. In OSTC, the assumption is relaxed such that the unlabeled examples can belong to both $C^{kk}$ and $C^{uu}$, making the problem open. Since samples from $C^{uu}$ are available during training, we consider that OSTC methods have a clear advantage over open-set learning methods, even though the available samples are unlabeled.

\citet{Chen-ACL-2024} use adversarial disagreement maximization to increase the difference between in-distribution and OOD examples, improving on their previous solution \citep{Chen-KDD-2023}. \citet{Kim-BigData-2022} propose a harder variant of OSTC, that is zero-shot. In this new setup, there are no labeled examples at all. The training data consists of a set of $n$ unlabeled texts $x_1, ..., x_n$ that must be classified into $k$ classes $c_1 ,..., c_k$. Each class $c_i$ is only specified through a set of words. The proposed method obtains better results than the corresponding closed-set solutions \citep{Meng-EMNLP-2020,Wang-ACL-2021}, when tested in an open-set setting.
LLMs have not been extensively studied for OSTC. \citet{Chen-ACL-2024} are among the few to compare small vs.~large language models on OOD detection and OSTC. They found that LLaMA2-7B performs poorly on both tasks, in both zero-shot and few-shot scenarios. Their study shows that prompting LLMs is not sufficient to accurately address OOD detection and OSTC, since a customized approach based on a smaller model can easily surpass zero-shot and few-shot LLM prompting.

\noindent
\textbf{Open-set learning and discovery.}
Ideally, during inference, an open-set system should be able to recognize patterns in the open data and dynamically create new relevant classes that would complement the set of classes $C^{kk}$ provided at train time. A solution to this problem should involve: (i) an \emph{open-set text classification} model to initially split the corpus into $C^{kk}$ and $C^{uu}$, (ii) a \emph{topic mining} method to identify new relevant topics in the open space, and (iii) a \emph{zero-shot learning} model to classify samples into the newly found topics. A further challenge is posed by the fact that the new classes should be semantically consistent with the existing classes. For example, if the existing classes are ``sports'', ``politics'' and ``economics'', a new class called ``culture'' is compatible, but one called ``breaking news'' is not. To obtain semantically equivalent classes, a knowledge base, such as WordNet, might come in handy. A possible architecture for the open-set learning and discovery setup is illustrated in Figure \ref{fig:OSD-Arch}. To the best of our knowledge, there is no framework for open-set learning and discovery in the field of text classification.

In the context of open-set text classification and discovery, instruction models can be used to perform the open-set text classification. However, it is difficult to discover new classes via prompting due to the large quantity of text and the relatively short-term memory of LLMs. However, using a reasoning model to interactively refine a list of possible topics is a path worth exploring. 

\vspace{-0.15cm}
\section{Comparison of Learning Frameworks}
\vspace{-0.15cm}

In Table \ref{tab:comparison}, we compare the three learning frameworks in terms of several key characteristics. From the comparative analysis, we next derive the main limitations of each learning framework.

\noindent
\textbf{Learning with background class.} The performance depends heavily on the quality and diversity of background samples. Since it is impossible to sample from UUCs, the background data will never fully represent the real open-world. Hence, UUCs will likely be treated as one of the KKCs.

\noindent
\textbf{Zero-shot learning.} Zero-shot models are sensitive to label wording, an aspect that is important to build representative class prototypes for UKCs. Without explicit supervision, the model might struggle to make subtle distinctions between fine-grained UKCs. UUC samples must map to one of the supplied classes, so zero-shot models cannot handle UUCs.

\noindent
\textbf{Open-set learning and discovery.} The learning problem is hard. It is difficult to define boundaries around KKCs, without examples of UUCs. It is even more difficult to classify UUCs without labeled samples. Moreover, it is hard to solve open-set learning and discovery via a single objective, requiring multiple optimization stages.

\begin{table}[t]
\centering
\setlength\tabcolsep{0.02em}
\small{
\begin{tabular}{lccc}
\toprule
\textbf{Characteristic} & \rotatebox[origin=c]{18}{\textbf{Universum}} & \rotatebox[origin=c]{18}{\textbf{Zero-shot}} & \rotatebox[origin=c]{18}{\textbf{Open-set}} \\
\midrule
Can detect OOD data            & \cmark & \xmark & \cmark \\
Set of classes is unbounded   & \xmark & \xmark & \cmark \\
Can classify OOD data         & \xmark & \cmark & \cmark \\
OOD training data required    & \cmark & \xmark & \xmark \\
Open-world assumption         & \xmark & \xmark & \cmark \\
Easy to optimize              & \cmark & \cmark & \xmark \\
\bottomrule
\end{tabular}
}
\vspace{-0.25cm}
\caption{Comparison of learning frameworks in terms of various characteristics.}
\label{tab:comparison}
\vspace{-0.4cm}
\end{table}



\vspace{-0.15cm}
\section{Conclusion and Future Work}
\vspace{-0.15cm}

\noindent
\textbf{Conclusion.}
Our survey examined three principal paradigms for text classification under shifting class distributions: learning with background class, zero-shot classification and open-set classification.

Learning with background class relies on auxiliary data to reject out-of-distribution examples. However, its effectiveness is highly sensitive to the quality and distinctiveness of the available KUC samples. When these auxiliary examples fail to capture real-life diversity, the approach struggles to generalize beyond its training set.
Zero-shot classification offers flexibility by leveraging semantic representations to extend recognition to unseen classes, yet it is inherently dependent on the robustness of the available semantic knowledge. In practice, when emerging classes deviate significantly from established semantic cues, zero-shot methods can falter, leading to misclassifications or an inability to capture nuanced differences between similar topics. Open-set classification, on the other hand, excels at detecting novelty through effective outlier identification. However, it typically lacks mechanisms for assigning meaningful labels or integrating these outliers into an existing classification framework. These methodologies have advanced our understanding of handling distribution shifts in text classification, but the field still lacks a unified framework capable of simultaneously discovering, categorizing, and adapting to emerging classes. A promising direction lies in continual learning approaches that can integrate the strengths of all three paradigms, while mitigating their individual weaknesses. 

\noindent
\textbf{Future directions.} Continual learning is likely one of the most promising paths towards unlocking open-set learning and discovery in the text domain. However, adapting continual learning techniques to mitigate open-set learning and discovery requires addressing additional complexities, such as novelty detection, adaptive knowledge retention, and dynamic model scalability. \citet{Liu-AI-2023} identify several challenges in the development of an open-word continuous learning system, such as autonomous novelty detection, automatic acquisition of labeled data via interaction, and risk assessment during self-initiated adaptation. One promising direction is disentanglement-based regularization, which separates learned representations into two spaces: a stable and generic one, and a flexible and task-specific one. This method enhances knowledge retention without restricting the adaptability of the model to new tasks, as demonstrated by \citet{Yufan-ACL-2021}. Another important avenue is optimizing memory selection for replay, where approaches such as k-means clustering can be used to retain only the most representative examples from previous tasks.

Research on continual learning for natural language generation (NLG) has also provided insights into how adaptive architectures can support evolving textual domains. \citet{Peng-ACL-2022} propose a transformer calibration mechanism to minimize interference between tasks in continual learning scenarios for NLG. This technique leverages attention calibration and feature recalibration to enable language models to incrementally adapt to new tasks, while preserving previously learned knowledge. Similar strategies could be beneficial for open-set continual learning in text classification, where models must adjust their representation space dynamically as new topics emerge.

\citet{Sun-ICLR-2020} introduce {LAMOL} (Language Modeling for Lifelong Language Learning), which reformulates various NLP tasks into language modeling. Under this framework, the model generates the replay samples from previously learned tasks, reducing catastrophic forgetting \citep{Kirkpatrick-PNAS-2017} without storing large historical datasets. Although originally designed for a closed set of tasks, LAMOL could be extended to open-set scenarios by incorporating a novelty detection module. Whenever the system encounters anomalous samples or unfamiliar topics, it can trigger an incremental update process that treats the new categories or subject area as separate tasks. This setup would allow the model to enlarge its internal knowledge base, maintaining prior knowledge, while adapting dynamically to unseen class distributions.

Drawing inspiration from CoLeCLIP \citep{Li-Arxiv-2024}, which introduces open-domain continual learning in the image domain through a combination of task-specific prompting and joint vocabulary learning, we can apply similar principles for text classification in open-set scenarios. The authors propose the development of shared representations and an adaptive prompting mechanism, which jointly enable a model to adjust to new tasks and classes as they emerge, while preserving previously acquired knowledge. Applied to text, this approach translates into designing dynamic prompts and an extensible vocabulary that guide classification under shifting class distributions, simultaneously detecting the emergence of new or unknown topics. For instance, building on the techniques employed by CoLeCLIP, a language model could learn task-specific prompts for different topics, continually adapting and expanding its internal vocabulary as it encounters new terms or concepts.  Such an approach would likely need to involve a regularization mechanism to prevent catastrophic forgetting, employing selective memory strategies to retain critical information from past data. 

While current open-set methods exhibit difficulties in threshold selection for open-space delimitation, LLMs offer significant potential for addressing class distribution shift. For instance, LLMs can provide auxiliary confidence signals by generating natural language explanations for classification decisions. When an LLM shows high uncertainty or produces inconsistent explanations for a sample, this signals potential inclusion into an unknown class, complementing traditional outlier detection methods. LLMs can also generate semantically coherent labels for discovered clusters (Figure \ref{fig:OSD-Arch}, step 4), ensuring consistency with existing classes, e.g.~proposing ``sports'' rather than ``tennis'' when existing classes are ``culture'', ``politics'' and ``economy''. 

\vspace{-0.15cm}
\section*{Acknowledgments}
\vspace{-0.15cm}

This work was supported by a grant of the Ministry of Research, Innovation and Digitization, CCCDI - UEFISCDI, project number PN-IV-P6-6.3-SOL-2024-0090, within PNCDI IV. This research is also supported by the project ``Romanian Hub for Artificial Intelligence - HRIA'', Smart Growth, Digitization and Financial Instruments Program, 2021-2027, MySMIS no.~351416.


\section*{Limitations}

To the best of our knowledge, this work covers the primary directions pertinent to the task of open-set text classification. However, some relevant studies may have been inadvertently missed due to limited visibility or other constraints.

\bibliography{references}

\appendix
\section{Appendix}

In the supplementary, we discuss related work on continual learning from text and how this research direction is useful in text classification under class distribution shift (Section \ref{sec_continual}). We also review the datasets that are commonly used in zero-shot and open-set learning from text (Section \ref{sec_data}).

\subsection{Continual Learning from Text}
\label{sec_continual}

\noindent
\textbf{Formal definition.}
Continual learning (CL) is a paradigm in machine learning focused on enabling systems to continuously learn from a stream of data over time. Unlike traditional approaches that assume a static data distribution and rely on fixed datasets, CL addresses challenges such as retaining knowledge from previous tasks (avoiding catastrophic forgetting \citep{Kirkpatrick-PNAS-2017}) and adapting to new tasks without extensive retraining \citep{Parisi-NN-2019}. CL can be defined as the process of training a model on a sequence of tasks, where each task $t$ can be characterized by a distinct data distribution. At any point in time, the model receives a batch of training samples $D_{t,b} = \{(x,y) \mid x \in \mathcal{X}_{t,b}, y \in \mathcal{Y}_{t,b}\}$, where $\mathcal{X}_{t,b}$ represents a subset of input samples and $\mathcal{Y}_{t,b}$ are the corresponding (task-specific) labels. The task identity is given by $t \in \{1, ..., \infty\}$, while the batch index is represented by $b \in \{1,..., \infty \}$. Theoretically, the continual learning process continues indefinitely during operation, hence the indexes $t$ and $b$ can grow endlessly. A task $t$ is formally defined by its data distribution $P(\mathcal{X}_t, \mathcal{Y}_t)$ \citep{Wang-TPAMI-2024}. 

\noindent
\textbf{Trends and approaches.}
CL is a relevant framework for open-set learning, as it can address challenges related to the open-set scenarios, especially in tasks such as text classification by topic when the topic distribution changes. These challenges include handling new classes and adapting to an evolving data distribution. CL can help to discover and learn new topics during inference. Both \citet{Chen-ICML-2014} and \citet{Gupta-ICML-2020} propose frameworks that enable models to accumulate knowledge over time, leveraging past knowledge to improve topic extraction, while mitigating catastrophic forgetting. These approaches integrate lifelong knowledge retention and transfer, ensuring that previously learned topics inform future learning tasks. By incorporating prior knowledge, these models enhance topic coherence, particularly in sparse data scenarios, where limited labeled data can affect the quality of extracted topics.
\citet{Chen-ICML-2014} introduce the Lifelong Topic Model (LTM), which mines prior knowledge from past document collections and integrates it into new topic modeling tasks through probabilistic inference. Similarly, \citet{Gupta-ICML-2020} propose the Lifelong Neural Topic Modeling (LNTM) framework, which relies on neural networks and selective data augmentation to balance stability and plasticity. LTM is designed for multi-domain topic extraction, identifying topic structures across different domains, while LNTM is more suitable for continuous document streams, maintaining topic stability, while adapting to novel information. Moreover, LTM employs a knowledge-based topic modeling approach that refines topic quality through extracted knowledge sets, whereas LNTM integrates topic regularization and selective data augmentation to enhance adaptation, while preventing excessive forgetting. 

\noindent
\textbf{Open-world continual learning.} Open-world continual learning is a learning paradigm in which a model incrementally acquires new tasks and, at the same time, detects and integrates previously unseen classes. 
In their recent work, \citet{Li-Arxiv-2025} propose a framework called HoliTrans, which transfers knowledge from both known and unknown class samples. HoliTrans uses Nonlinear Random Projection to create clearer representations that help to separate new samples from previously learned classes. The method also relies on Distribution-Aware Prototypes, which dynamically adapt as new classes appear. These components help to better distinguish between new and previously seen classes, particularly when unknown classes repeatedly occur.

\subsection{Text Datasets for Class Distribution Shift}
\label{sec_data}

\begin{table*}[t!]
    \setlength\tabcolsep{0.2cm}
    \centering
    \begin{tabular}{llrrccc}
        \toprule
        \textbf{Dataset} & \textbf{Task} & \rotatebox[origin=c]{90}{\textbf{\#samples}} & \rotatebox[origin=c]{90}{\textbf{\#classes}} & \rotatebox[origin=c]{90}{\textbf{Zero-shot}} & \rotatebox[origin=c]{90}{\textbf{Open-set}} & \rotatebox[origin=c]{90}{\textbf{Continual}} \\
        \midrule
        AG News \citep{Zhang-NIPS-2016} & Topic categorization & 127k & 4 & \cmark & \cmark & \cmark \\
        DBPedia \citep{Lehmann-SW-2015} & Topic categorization & 630k & 14 & \cmark & \cmark  & \cmark\\
        20 Newsgroup \citep{Lang-ICML-1995} & Topic categorization & 18k & 20 & \cmark & \cmark & \xmark \\
        Yahoo! Answers \citep{Zhang-NIPS-2016} & Topic categorization & 1.4m & 10 & \cmark & \cmark & \xmark \\
        SST-2 \citep{Socher-EMNLP-2013} & Polarity classification & 69k & 2 & \cmark & \xmark & \xmark \\
        IMDb \citep{Maas-ACL-2011} & Polarity classification & 50k & 2 & \cmark & \xmark & \xmark \\
        Amazon Product Reviews \citep{Chen-CIKM-2013} & Topic modeling   &50k  & 50 & \xmark & \cmark & \cmark \\
        TMNtitle \citep{Gupta-ICML-2020}  & Topic modeling   &32.6k  & 7 & \xmark & \xmark & \cmark \\
        R21578title \citep{Gupta-ICML-2020} & Topic modeling   &10.8k & 90 & \xmark & \xmark & \cmark \\

        \bottomrule
    \end{tabular}
        \caption{List of datasets that are commonly used in text classification under class distribution shift. For the selection of datasets, we report some of their basic statistics, as well as the tasks where they are commonly employed.}
    \label{tab:datasets}
\end{table*}

In Table \ref{tab:datasets}, we provide a list of text classification datasets that are often used in text classification under class distribution shift. These datasets are typically configured for a standard supervised classification setup. We further explain how the listed datasets are modified to serve the zero-shot, open-set, and continual learning setups, respectively.

\noindent
\textbf{Text datasets for zero-shot classification.}
A possible methodology for zero-shot testing is proposed by \citet{Yin-EMNLP-2019}, alongside a few suitable datasets.  For topic categorization, the authors propose the Yahoo!~Answers \citep{Zhang-NIPS-2016} dataset. This dataset is reorganized into \emph{dev} and \emph{test} splits, both containing all 10 labels. The \emph{dev} split consists of 6k instances per label, while the \emph{test} split has 10k instances per label. For the labels-partially-unseen scenario, two variants of the dataset are created, with different partitions of seen/unseen classes, both balanced. For labels-fully-unseen there is no training set. 
\citet{Sanh-ICLR-2022} employ AG News \citep{Zhang-NIPS-2016} and DBPedia \citep{Lehmann-SW-2015} for zero-shot topic categorization, and IMDb \citep{Maas-ACL-2011} for polarity classification. There is no training set for the target task. Similarly, \citet{Meng-EMNLP-2020} use the AG News, DBPedia, IMDb and Amazon Product Reviews \citep{Chen-CIKM-2013} datasets, with all the labels removed. In general, we observe that the preferred datasets are large scale.

\noindent
\textbf{Text datasets for open-set classification.}
The common way to test an open-set solution is to reorganize a dataset designed for supervised text classification. To this end, some classes are reserved to play the role of the open space and are not used during training. While this scenario may seem similar to the zero-shot labels-partially-unseen setup, the labels for the reserved classes are not used. During testing, the system must simply detect them as belonging to the open space, without classifying them into classes. Many works perform multiple experiments, varying the KKCs-to-UUCs ratio between 20\% and 100\%. It is worth noting that a percentage of 100\% known classes is not a closed-set problem, because the model can still classify instances as open space.

\citet{Kim-BigData-2022} report experiments on AG News and DBPedia. The KKCs-to-UUCs ratio is chosen among three values: 25\%, 50\% and 75\%. For AG News, all the combinations of known / unknown classes having at least one known and one unknown class are compared. Some studies \citep{Chen-KDD-2023,Chen-ACL-2024} conduct experiments on three datasets, namely AG News, DBPedia and Yahoo! Answers, reserving two, four or six classes to play the role of UUCs. Other papers \citep{Walkowiak-ASMBI-2018,Walkowiak-ICAISC-2019} use custom datasets, specifically tailored for the open-set task. \citet{Walkowiak-ICAISC-2019} are the only ones who use data in a foreign language, namely Polish. 

\noindent
\textbf{Text datasets for continual learning.}
\citet{Gupta-ICML-2020} report experiments on Amazon Product Reviews  \citep{Chen-CIKM-2013}, 20NSshort, TMNtitle and R21578title datasets. They use these datasets as future tasks in a continual learning framework for topic modeling, testing the ability of models to learn from short texts, retain knowledge, and adapt to new domains without catastrophic forgetting. The Amazon Product Reviews dataset is preprocessed according to the methodology described by \citet{Chen-CIKM-2013}. The preprocessing includes sentence detection, lemmatization, and POS tagging, as well as the removal of punctuation, stop words, and rare terms. To avoid excessive thematic overlap, the domain name of each collection is removed. 20NSshort is a subset of 20 Newsgroups \citep{Lang-ICML-1995}, containing only documents with fewer than 20 words, making topic inference challenging due to sparse content. Similarly, R21578title is derived from Reuters-21578 \citep{Lewis-UCI-1987}, retaining only article titles, further limiting contextual information. TMNtitle comes from Tag My News (TMN) \citep{Vitale-ECIR-2012} and includes only news headlines, reflecting real-world short-text classification tasks.


\citet{Sun-ICLR-2020} use DBPedia as part of a sequential continual learning setup alongside other datasets, such as AG News, Amazon Product Reviews, and Yahoo!~Answers. As the authors sequentially introduce new classification tasks, DBPedia helps them to assess how well the model could retain previously learned categories, while adapting to new ones.

\noindent
\textbf{General remark.} We find that datasets used in literature do not always reflect real-world open-set scenarios. This is mostly due to the fact that nearly all datasets were introduced more than 10 years ago. We thus believe that collecting new resources that cover realistic open-set setups is an important avenue for future research.

\end{document}